\documentclass[lettersize,journal]{IEEEtran}
\usepackage{amsmath,amsfonts}
\usepackage{algorithmic}
\usepackage{algorithm}
\usepackage{array}
\usepackage[caption=false,font=normalsize,labelfont=sf,textfont=sf]{subfig}
\usepackage{textcomp}
\usepackage{stfloats}
\usepackage{url}
\usepackage{verbatim}
\usepackage{graphicx}
\usepackage{cite}
\usepackage{pifont}
\usepackage{booktabs}       
\usepackage{amsfonts}       
\usepackage{nicefrac}       
\usepackage{microtype}      
\usepackage{xcolor}         
\usepackage{array,multirow}
\usepackage{float}
\usepackage{graphicx}
\usepackage{subfig}
\usepackage{amsmath}
\usepackage{wrapfig}
\usepackage{color}

\begin{document}

\title{AttMOT: Improving Multiple-Object Tracking by Introducing Auxiliary Pedestrian Attributes}

\author{Yunhao Li, Zhen Xiao, Lin Yang, Dan Meng, Xin Zhou, Heng Fan, Libo Zhang$^\dag$
\thanks{$^\dag$Corresponding author(libo@iscas.ac.cn).}
\thanks{Yunhao Li, Zhen~Xiao, Xin Zhou and Libo~Zhang are with the Department of State Key Laboratory of Computer Science, Institute of Software Chinese Academy of Science.}
\thanks{Lin Yang is with Turing Quantum Company. Dan Meng is with OPPO Research Institute.}
\thanks{Heng~Fan is with the Department of Computer Science and Engineering, University of North Texas, Denton, USA}
}


\maketitle

\begin{abstract}
Multi-object tracking (MOT) is a fundamental problem in computer vision with numerous applications, such as intelligent surveillance and automated driving. Despite the significant progress made in MOT, pedestrian attributes, such as gender, hairstyle, body shape, and clothing features, which contain rich and high-level information, have been less explored. To address this gap, we propose a simple, effective, and generic method to predict pedestrian attributes to support general Re-ID embedding. We first introduce AttMOT, a large, highly enriched synthetic dataset for pedestrian tracking, containing over 80k frames and 6 million pedestrian IDs with different time, weather conditions, and scenarios. To the best of our knowledge, AttMOT is the first MOT dataset with semantic attributes. Subsequently, we explore different approaches to fuse Re-ID embedding and pedestrian attributes, including attention mechanisms, which we hope will stimulate the development of attribute-assisted MOT. The proposed method AAM demonstrates its effectiveness and generality on several representative pedestrian multi-object tracking benchmarks, including MOT17 and MOT20, through experiments on the AttMOT dataset. When applied to state-of-the-art trackers, AAM achieves consistent improvements in MOTA, HOTA, AssA, IDs, and IDF1 scores. For instance, on MOT17, the proposed method yields a +1.1 MOTA, +1.7 HOTA, and +1.8 IDF1 improvement when used with FairMOT. To encourage further research on attribute-assisted MOT, we will release the AttMOT dataset. 
\end{abstract}


\section{Introduction}
\IEEEPARstart{M}{ulti-object} tracking (MOT) is an important and fundamental task in computer vision~\cite{bewley2016simple, wojke2017simple, wang2020towards, zhang2022bytetrack}, which aims at estimating the bounding boxes and identities of objects in videos. This task has been deployed in various applications including intelligent monitoring, autonomous driving, video analysis, and human activity recognition~\cite{wang2013approach, luo2017learning}, etc.

\begin{figure}[t]
	\centering
	\vspace{2mm}
	\includegraphics[width=0.9\linewidth]{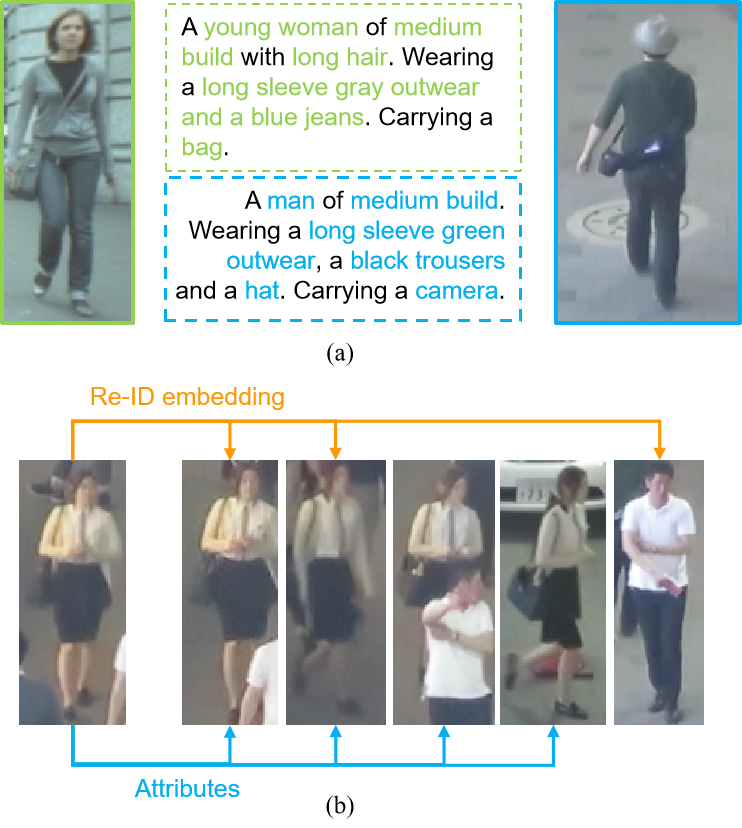}
	\vspace{-1mm}
	\caption{(a) Examples of how attributes describe a person. (b) Advantages of using attributes. It's easy to use attributes to describe a person, and when with enough attributes and limited pedestrians we can almost uniquely identify an id. Another benefit of using attributes to associate targets is that attributes are motion and occlusion irrelevant while regular Re-ID embedding is not.}
	\label{fig:0}
	\vspace{-1mm}
\end{figure}

Broadly speaking, most of the existing MOT methods contain two components, i.e, detection and association. Early approaches such as~\cite{wojke2017simple, mahmoudi2019multi, fang2018recurrent} usually process these two components separately: a detection model first detects objects of interest and then an association model extracts re-identification (Re-ID) features or utilizes motion information IoU to associate the predicted targets between frames. Recent methods like~\cite{zhang2021fairmot, zhang2022bytetrack, zhou2020tracking, liang2022rethinking} usually integrate these two components into a single model. Hence, current advances in MOT can be roughly divided into two categories, i.e, trying to get better detection results or trying to improve the association of the detected targets. There has been remarkable progress in detection~\cite{ren2015faster, he2017mask, zhou2019objects, fu2020model} and association~\cite{zheng2017person, chen2018real, zhang2022bytetrack} respectively, which greatly boosts the overall tracking accuracy. Currently, the MOT community mainly focuses on the association part, in which the problem of how to extract discriminative pedestrian features is of great interest, yet the area of attribute-assisted MOT remains under-explored. Pedestrian attributes provide detailed and significant high-level information about a person, including \textit{gender, body shape, the color of clothes, and so on}. Figure ~\ref{fig:0}(a) shows two examples of how attributes are used to describe a person in the MOT17 dataset. Besides, pedestrian attributes are entirely motion irrelevant and occlusion resistant, e.g. as shown in Figure~\ref{fig:0}(b), using regular Re-ID embedding only may result in missing correct matches (miss matching the third image on the right because of white occlusion) and arising incorrect matches (incorrectly matches the last image on the right because they have similar motion and composition). 

AttMOT does not specifically focus on occlusion, and even when a pedestrian is partially occluded, their attributes still exist. For example, the women in the fourth image in Figure~\ref{fig:0}(b) still has the attribute \textit{black skirt}. However, if a pedestrian's attribute is located behind them and the camera angle is from the front, this attribute is disregarded. This decision is made because, for the purpose of pedestrian tracking, we aim to ensure that each pedestrian's attributes remained constant throughout the tracking process. Therefore, we aim to train a feature extractor that could resist occlusion to some extent. This necessitate annotating specific attributes in the dataset even when occlusion occurred. It is worth noting that the attributes we select are relatively easy to observe, which means that complete occlusion is less likely to occur. This further supports the efficacy and practicality of our dataset for pedestrian tracking with attribute annotations.

We posit that attribute information can serve as a valuable complement to general Re-ID embedding. In this paper, we investigate the potential of utilizing these attributes as a robust reinforcement for existing MOT trackers.

\begin{figure*}[t]
	\centering
	\includegraphics[width=0.98\linewidth]{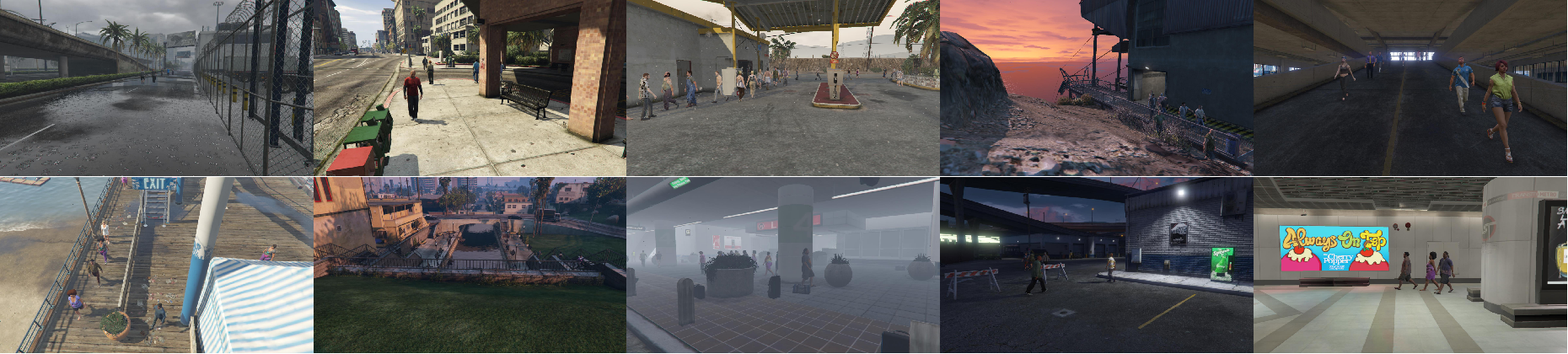}
	\caption{Dataset exhibition. We propose a large and highly diverse pedestrian tracking dataset that provides not only identity and location information but also semantic attributes. We use such a dataset to train a pedestrian attribute recognition module and attribute-assisted trackers.}
	\label{fig:1}
\end{figure*}

To achieve this goal, the key challenge is to design a model that makes good use of attributes as a complementary component and to build a method for training such a model, and to our understanding, the latter is the more tremendous obstacle. There are several attempts to assist person re-identification (Re-ID) with attributes, some works~\cite{su2017multi, wang2018transferable, shi2020person} regarded attributes as discriminative features to enhance or even replace regular Re-ID features, while the others~\cite{lin2019improving, han2018attribute, ling2019improving, zhang2020person, tay2019aanet} use attributes to help co-training. This raises a question: \textit{Why is attribute still less used in MOT while attribute-assisted or enhanced person Re-ID is frequently discussed?} One of the most important reasons is datasets. To be precise, as far as we know, no public pedestrian tracking dataset with semantic attribute annotation is available now. Meanwhile, it is a truism that MOT approaches are extremely data-hungry now, current deep neural networks usually require hundreds of thousands of images and annotations to learn robust and meaningful feature representation, not to mention Transformer-based methods~\cite{carion2020end, zhu2020deformable, ma2017transt, sun2020transtrack, meinhardt2022trackformer}. However, real-world data collection and labeling are annoyingly time-consuming and expensive, let alone the potential privacy problems. Privacy obtain increasing attention and regulation has been seriously concerned, e.g. European Union passed General Data Protection Regulations (GDPR~\cite{voigt2017eu}) to protect citizen privacy. To address the above issues, in this paper we present our AttMOT, a synthetic dataset for pedestrian tracking and Re-ID with highly accurate and rich labels of ids, 2D bounding box coordinates, and semantic attributes. As previously mentioned, this is the first time a pedestrian tracking dataset with attribute annotations has been made available. We used a novel data generation process utilizing virtual games to create the dataset. With mature game technology, real-world simulations provide a powerful way to generate varied data under different conditions, such as various weather conditions, viewpoints, pedestrian identities, and ambient lighting, at a low cost.Some images from AttMOT is displayed in Figure~\ref{fig:1}, and we will give a detailed introduction in Chapter~\ref{Dataset}.

Another issue that poses a challenge to our design is how to incorporate pedestrian attributes into MOT trackers. In this paper, we propose the Attribute-Assisted Method (AAM) which leverages both Re-ID embedding and attributes for MOT. As discussed earlier, current trackers can be classified into two groups based on whether they employ a joint model to detect and associate targets or not (two-step and one-step approaches). Therefore, we develop different assistance methods for different types of trackers. For instance, for approaches that use separate detection and association models, we employ a simple network to extract pedestrian attributes and build a vector. The attribute vector is then concatenate with the original Re-ID embedding to create a new attribute-assisted feature. Although this method is relatively straightforward, our experiments have demonstrated its effectiveness. As for one-step trackers, we propose a method that uses an attention mechanism~\cite{vaswani2017attention} to simultaneously extract Re-ID embedding and attribute vector. We evaluate the performance of our proposed AAM on two representative datasets of MOTChallenge, i.e. MOT17~\cite{milan2016mot16} and MOT20~\cite{dendorfer2020mot20}, and the results suggest that our method can improve the performance of state-of-the-art approaches.

Our contributions are summarized as: 
\begin{itemize}
	\item
	We propose a synthetic dataset with semantic attribute annotations, which contains 450 sequences and more than 6.6 million precisely annotated pedestrian instances.
	\item
	To enhance existing MOT trackers, we design a simple but effective method called AAM, which utilizes attribute information.
	\item
	By evaluating the proposed approach and dataset on several MOT benchmarks for various tasks such as pedestrian recognition, categorization, and tracking, we validate the efficacy of attributes-assisted MOT.
\end{itemize}

\section{Related Work}
\subsection{Multi-object tracking (MOT)}
MOT, along with its variants like monocular multi-object tracking~\cite{liu2014learning} and tiny object tracking~\cite{zhu2023tiny}, have always been a very important and fundamental kind of problem in computer vision, and it is widely used in many domains such as intelligent surveillance and automatic driving. In this subsection, we'd like to review some representative trackers. We divide the existing works into two categories based on their design paradigm.

 The first type of tracker follows the tracking-by-detection (TBD) paradigm, which views MOT as a two-step task. In the first step, a target detection algorithm is used to detect the targets of interest in each frame and obtain the corresponding information such as coordinates and confidence score. Focusing on achieving better detection results, some approaches have been introduced~\cite{yu2016poi, ren2015faster, yang2016exploit, zhou2018online, mahmoudi2019multi, cai2018cascade}. The next step is correlating detection targets in different frames using different association methods. Most of the existing work focuses on this part of the problem, which can be broadly classified into motion cue-based methods and appearance cue-based methods. Research on motion cues-based methods roughly starts with SORT~\cite{bewley2016simple} and IOU-Tracker~\cite{bochinski2017high}. SORT first uses Kalman Filter~\cite{kalman1960new} to predict the future locations of the tracklets and uses the Hungarian algorithm~\cite{kuhn1955hungarian} to match detected targets and tracklets. IOU-Tracker directly calculates the overlap area of the trajectories and detection in adjacent frames for the association. Both SORT and IoU-Tracker are very widely used, mainly because they are simple and fast. Further studies~\cite{zhang2020long, han2022mat, zhang2022bytetrack} on this method focus on the problem of possible failure in challenging cases of crowded scenes and fast motion. Location and motion similarity is accurate in short-range matching, and appearance similarity is helpful in long-range matching. There are some works~\cite{bae2014robust, tang2017multiple, sadeghian2017tracking, chen2018real, fang2018recurrent} trying to get better and enhanced features either.

Another MOT architecture called Joint-Detection-and-Tracking, which jointly detects and tracks objects using a single network and completes them in a single step, has recently attracted more research attention. We follow the classification strategy of Zhang et al.~\cite{zhang2021fairmot} and broadly divide trackers using this framework into two categories, one is joint detection and Re-ID~\cite{voigtlaender2019mots, wang2020towards, liang2022rethinking, lu2020retinatrack}, and the other is joint detection and motion prediction~\cite{feichtenhofer2017detect, zhou2020tracking,  peng2020chained, sun2020transtrack}.The first type of trackers are often based on great detectors, e.g. Track-RCNN~\cite{voigtlaender2019mots} adds Re-ID head to Mask-RCNN~\cite{he2017mask}, and JDE~\cite{wang2020towards} is built on top of YOLOv3~\cite{redmon2018yolov3}. The latter class of methods learns detection and motion features in a single network, representative trackers including Tracktor~\cite{sridhar2019tracktor}, Chained-Tracker~\cite{peng2020chained}, and CenterTrack~\cite{zhou2020tracking}. It is also worth mentioning that with the spread of Transformer~\cite{han2021transformer} in computer vision, the attention mechanism~\cite{vaswani2017attention} is now widely used in multi-object tracking and its sub-tasks like person Re-ID~\cite{zhou2021attention}.

\subsection{Synthetic datasets}
Current deep networks are extremely data-hungry, and they usually require hundreds of thousands of images and annotations to learn robust and meaningful feature representations. Unfortunately, manual data collection is both expensive and time-consuming. Meanwhile, large real-world datasets are also vulnerable to privacy issues. As a possible solution, synthetic datasets are introduced to many computer vision tasks. For example, PersonX~\cite{yang2004finding} and SyRI~\cite{bak2018domain} are used for person Re-ID, GCC Dataset~\cite{wang2019learning} is used for human counting, JTA Dataset~\cite{fabbri2018learning} is used for pedestrian pose estimation, Virtual KITTI~\cite{gaidon2016virtual}, CARLA~\cite{dosovitskiy2017carla} and Europilot~\cite{richter2016playing} are applied in autopilot training, SYNTHIA~\cite{ros2016synthia} and Playing for Data~\cite{richter2016playing} are used for semantic segmentation, and MOTSynth~\cite{fabbri2021motsynth} is designed for multiple tasks.

Using synthetic data in computer vision allows us to easily generate massive amounts of data in a way that is cheaper, faster, and more accurate than real-world data collection. We can also generate data that is difficult to collect in the real world, such as images of traffic conflict zones. Based on some of the existing works, we observe that using synthetic data can retain data value while eliminating sensitive information to some extent~\cite{gaidon2016virtual, kar2019meta}, and we can simulate scenarios almost entirely according to our needs~\cite{hattori2015learning}. But synthetic data also has its drawbacks. One major problem is that there is a non-negligible domain gap between synthetic data and real-world data. We can use methods like domain adaptation~\cite{deng2018image} and data generation~\cite{zheng2019joint} to alleviate the problem. Besides, as the experiments in ~\cite{fabbri2021motsynth} shows, the diversity of video sequences has a significant effect on bridging the gap between real and virtual data, and this is relatively easy to achieve in virtual games. Real-world data often has many limitations, such as high collection costs and difficulty in obtaining, which leads to a lack of quantity and diversity in data. In contrast, in virtual games, it is relatively easy to generate large amounts of data while controlling scenes and environments to increase data diversity. By increasing the diversity of video sequences in virtual games, models can better learn and understand different scenes and environments. For example, in the field of autonomous driving, virtual games can generate different road scenes and weather conditions, allowing the model to better adapt to various driving scenarios. In addition, in the field of video analysis, virtual games can generate different characters, objects, and backgrounds, helping models better recognize and understand different video content.

\subsection{Attribute-assisted Person Re-ID}
In some early attempts, pedestrian attribute recognition~\cite{wang2022pedestrian} is introduced to person Re-ID due to its natural similarity between pedestrian attribute learning and feature extraction. Person attributes can be either used as an effective complement to identity features~\cite{layne2012person, li2014clothing, su2017multi, wang2018transferable, li2019attribute, liu2017video} or considered as supervision to help co-training~\cite{lin2019improving, ling2019improving, han2018attribute, zhang2020person, wang2019learning_2}. Recently, there have been some works about attribute-assisted video person Re-ID~\cite{chai2022video}, yet attribute-assisted MOT is still less studied. To this end, we propose a comprehensive study on attribute-assisted MOT and prove its validity.

\begin{figure*}[t]
	\centering
	\includegraphics[width=0.98\linewidth]{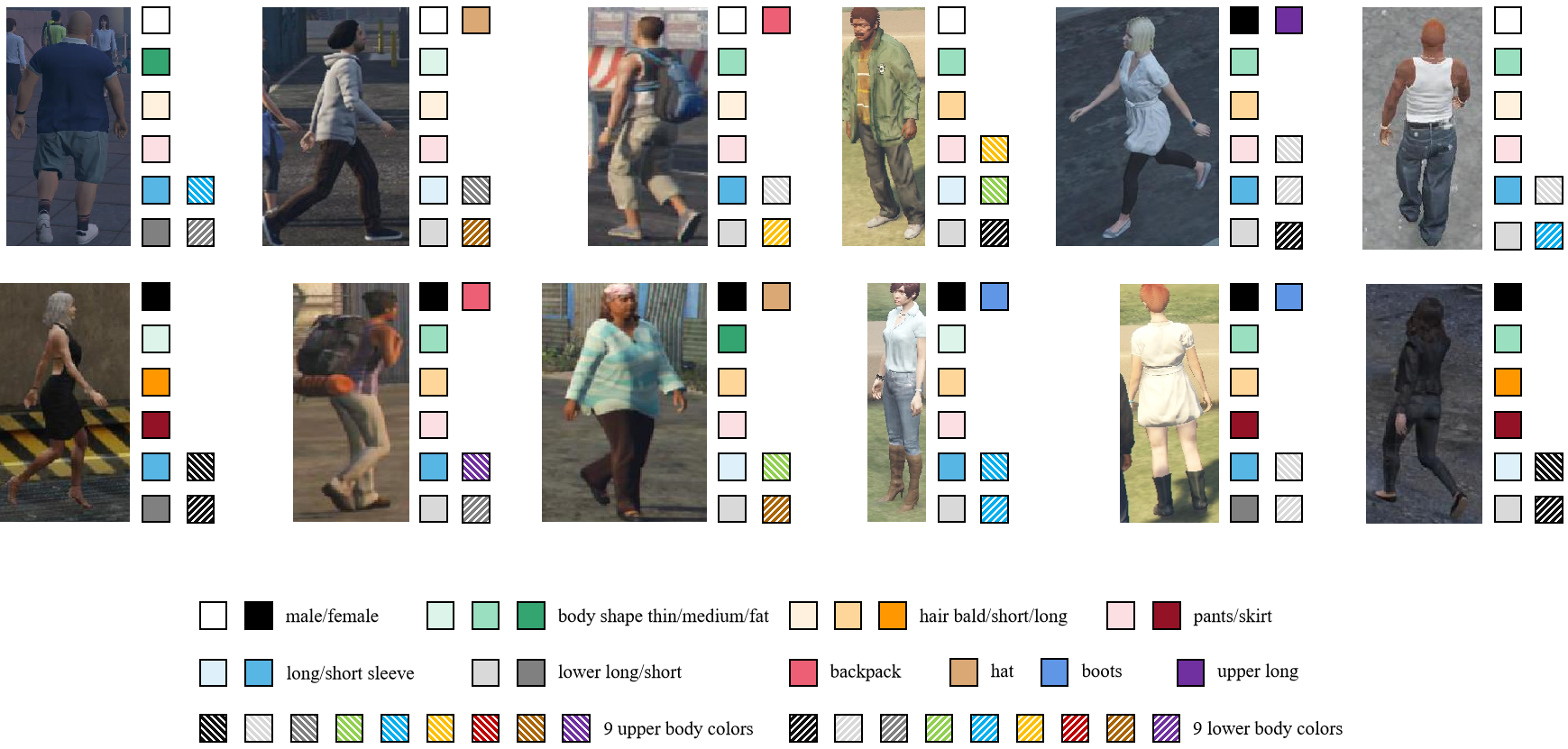}
	\caption{Pedestrian examples with attribute labels. Our AttMOT contains a rich set of pedestrian ids and attributes. In this image, we display several examples of pedestrians with some representative attributes.}
	\label{fig:2}
\end{figure*}

\section{Dataset} \label{Dataset}
In this section, we introduce our synthetic dataset, AttMOT. In first subsection we illustrate how to generate the dataset, including image extraction and label tagging, and then the selection of attributes is explained in the second subsection along with the reasons. Finally, a statistical analysis is presented on our dataset, and it is compared to other related real-world and synthetic datasets.

\subsection{Dataset Generation}
We follow the guidelines of prior works~\cite{richter2017playing, fabbri2021motsynth} about synthetic datasets and utilize the Script Hook V library to extract images and labels from Grand Theft Auto V (GTA-V). 

\subsubsection{Scene Settings}
GTA-V provides a virtual world of 130 $km^2$ (about an eighth of Los Angeles County), containing a large variety of scenarios. To this end, we manually explored almost the entire map of GTA-V and chose more than 50 scenarios including \textit{busy streets, ghettos, parks, parking lots, interior scenes, wharves, wasteland, and the like}. Some of the scenarios are shown in Figure~\ref{fig:1}, e.g, indoor scenes in the pictures on the far right, the city park scene in the second picture in the second row, and the mountain top in the fourth picture in the first row. After choosing the scenario, camera viewpoints are manually set and various pedestrians are randomlly generated. As shown in Figure~\ref{fig:1}, AttMOT is diverse in terms of perspective variations. Finally, in order to make the scene more realistic, the trajectories of each pedestrian are manually set and pedestrians from 160 models with a lot of different combinations of attributes are generated, e.g., ~\textit{clothes, hair, skin color, and masks}. 

\subsubsection{Screenplay Recording}
After finishing the creation of the scene, we then simulate virtual world dynamics and render different views of the simulated environments. In order to maximize the advantages of the virtual game GTA-V, we used a random time clock and weather conditions when recording. For the weather conditions, we followed prior work~\cite{fabbri2021motsynth} and used 9 kinds of weather: ~\textit{clear, extra sunny, cloudy, overcast, rainy, thunder, smog, foggy, and blizzard}. Some scenes with different weather conditions and times can be seen in Figure~\ref{fig:1}, e.g, the night scene in the fourth picture in the second row, heavy fog in the third picture in the second row, and the rainy scene in the first image. 

\begin{figure}[t]
	\centering
	\vspace{2mm}
	\includegraphics[width=0.98\linewidth]{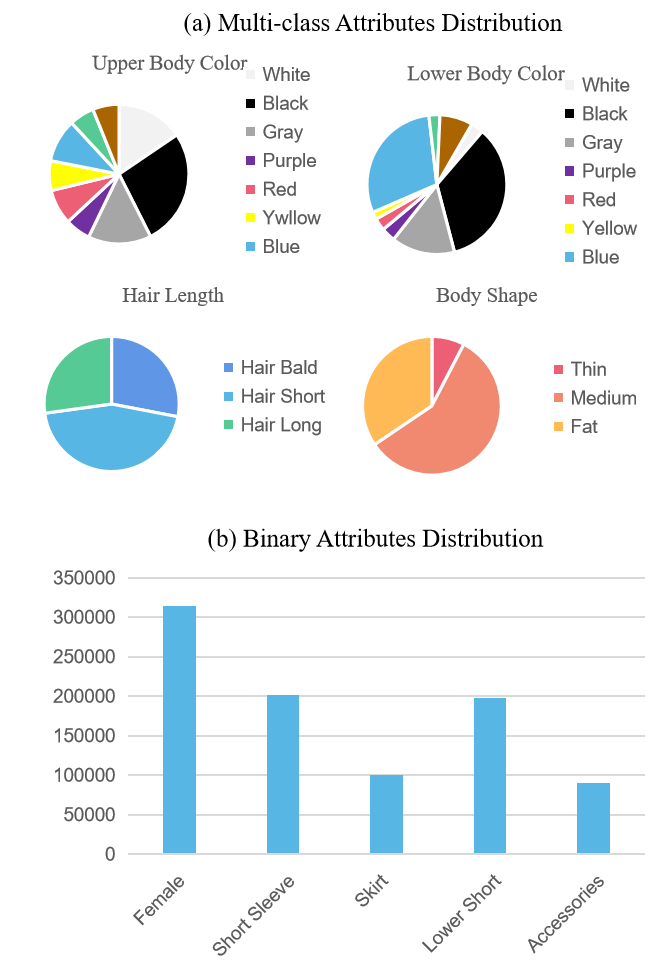}
	\vspace{-1mm}
	\caption{Distributions of various attributes including binary attributes and multi-class attributes. The pie charts display four multi-class attributes, i.e. upper body color, lower body color, hair length, and body shape. The histogram shows the distribution of binary attributes. 'Accessories' is the combination of several accessory attributes. i.e. hat, backpack, boots, and long upper clothes.}
	\label{fig:3}
	\vspace{-1mm}
\end{figure}

\subsubsection{Label Alignment}
 Common MOT datasets like MOT17~\cite{milan2016mot16} and MOT20~\cite{dendorfer2020mot20} only provide labels like 2D bounding box coordinates and pedestrian ids, and these annotations can be exported directly from GTA-V using the Script Hook V library. The semantic attributes discussed earlier are quite diverse and cannot be easily extracted. Manually labeling them on a frame-by-frame basis is not only time-consuming but also impractical. As a solution, we utilized OpenIV to access the model library of GTA-V and annotated attribute annotations in the unit of the model attribute. By combining these attributes, we were able to obtain a larger number of annotated pedestrians than those labeled manually.
\IEEEpubidadjcol

\begin{table}[t]
	\centering
	\renewcommand{\arraystretch}{1.1}
	\tabcolsep=2.8mm
	\caption{Detailed annotations in AttMOT.}
	\begin{tabular}{l|cc}
		\toprule[1.2pt]
		Class  &  Attributes \\
		\midrule[0.8pt]
		gender & male/female \\
        body shape & body thin/medium/fat \\
        hair length & hair bald/short/long \\
        upper body & long/short sleeve, upper-body long/short \\
        lower body & pants/skirt, lower-body long/short \\
        accessories & backpack, hat, boots \\
        color & 9 colors of lower-body/upper-body clothing \\
 	\bottomrule[1.2pt]
	\end{tabular}
	\label{tab:attrs}
\end{table}

\subsection{Attribute Annotations}
As a synthetic dataset, AttMOT contains a rich set of pedestrian ids and attributes. To make it more suitable for the MOT task, we choose 32 attributes, which are shown in Table~\ref{tab:attrs}. We display several pedestrian examples with some representative attributes in Figure~\ref{fig:2}, e.g. we can see a knitted hat in the second image of the first row, a backpack in the second row, and boots in the fourth and fifth images of the second row. Note that a pedestrian can have multiple colors of upper or lower body clothing, for example, the upper body color of the pedestrian in the fourth column of the first row can be green and yellow at the same time, as shown in Figure~\ref{fig:2}. 

Theoretically, a larger number of attributes beyond the 32 we utilized could be obtained, such as skin color, hair color, and shoe type. However, the selection of attributes was based on their relevance to the MOT task. Some attributes were deemed irrelevant in MOT due to various reasons. For example, small objects, such as earphones and necklaces, were almost imperceptible. Shoe color and type were deemed irrelevant as they were constantly obscured. Similarly, the skin color of pedestrians in tracking videos tended to be homogeneous. Consequently, after careful consideration of these factors, we opted for the chosen 32 attributes. It is widely acknowledged that the use of synthetic datasets can effectively mitigate the problem of imbalanced samples. We show the distributions of various binary and multi-class attributes that we use in Figure~\ref{fig:3}.

\subsection{Statistical Analysis}
\begin{figure*}[t]
	\centering
	\includegraphics[width=0.98\linewidth]{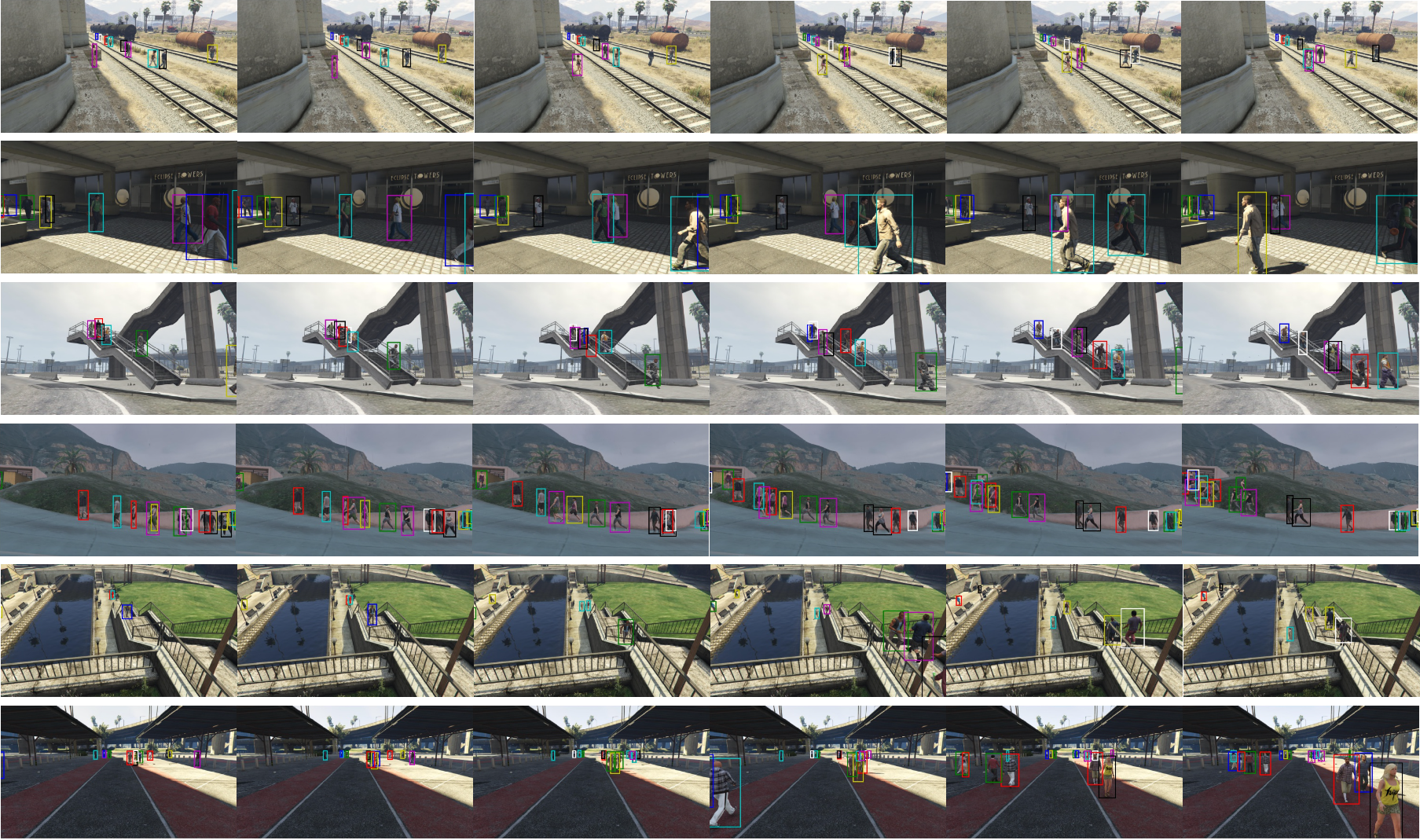}
	\caption{Exhibition of annotated samples. Visualization of annotated sample sequences with different time intervals: 30 frames (row 1-2), 60 frames (row 3-4), and 120 frames (row 5-6).}
	\label{fig:4}
\end{figure*}

In this subsection, we perform a statistical analysis of our proposed synthetic dataset. AttMOT consists of 450 sequences, 810k densely and accurately annotated frames and over 6.6 million bounding boxes. In particular, we demonstrate several annotated examples in Figure~\ref{fig:4}, from which we can see that our annotations are accurate and consistent. For regular annotations, the same annotation format with MOT17~\cite{milan2016mot16} and MOT20~\cite{dendorfer2020mot20} are used for each pedestrian target, and for attribute annotations, we simply use a 32-d binary vector to represent 32 attributes: 1 means that the pedestrian has the corresponding attributes, and 0 means the opposite. 

In addition, AttMOT is compared with some popular pedestrian tracking benchmarks, namely MOT17 and MOT20. From the comparison in Table~\ref{tab:1}, we can see that AttMOT is larger than regular pedestrian tracking datasets in terms of frames and the number of instances, and its scenes and pedestrian appearances are much more diverse and complex. Compared to other synthetic datasets such as JTA~\cite{fabbri2018learning} and GTA~\cite{krahenbuhl2018free}, we still maintain the advantage of the data scale (except for MOTSynth~\cite{fabbri2021motsynth}). Also, we'd like to reiterate that our AttMOT is the first pedestrian tracking dataset with attribute annotations.

\begin{table}[t]
	\centering
	\renewcommand{\arraystretch}{1.1}
	\tabcolsep=2.8mm
	\caption{Comparison with publicly available datasets for pedestrian tracking related tasks. AttMOT is the first attribute dataset of its kind, and we mainly compare the numbers of annotated frames and instances.}
	\begin{tabular}{l|cccc}
		\toprule[1.2pt]
		& Frames  &  Instances & Sequences & Attributes\\
		\midrule[0.8pt]
		MOT17~\cite{milan2016mot16} & 11.2k & 292k & 13 & \ding{56} \\
		MOT20~\cite{dendorfer2020mot20} & 13.4k & 1,652k & 4 & \ding{56}\\
            \midrule[0.8pt]
            GTA~\cite{krahenbuhl2018free} & 280k & 3,875k & - & \ding{56}\\
            JTA~\cite{fabbri2018learning} & 460k & 15,34k & 128 & \ding{56}\\
            MOTSynth~\cite{fabbri2021motsynth} & 1382k & 40,780k & 576 & \ding{56}\\
            \midrule[0.8pt]
            ~\textbf{AttMOT} & ~\textbf{810k} & ~\textbf{6,694k} & ~\textbf{450} & \ding{52}\\
 		\bottomrule[1.2pt]
	\end{tabular}
	\label{tab:1}
\end{table}

\section{Methodology}
In this section, two simple but effective baseline methods are proposed in the first subsection. And then we conduct an analysis and exploration of how to better merge regular Re-ID embeddings and attributes.

\begin{figure*}[t]
	\centering
	\includegraphics[width=0.9\linewidth]{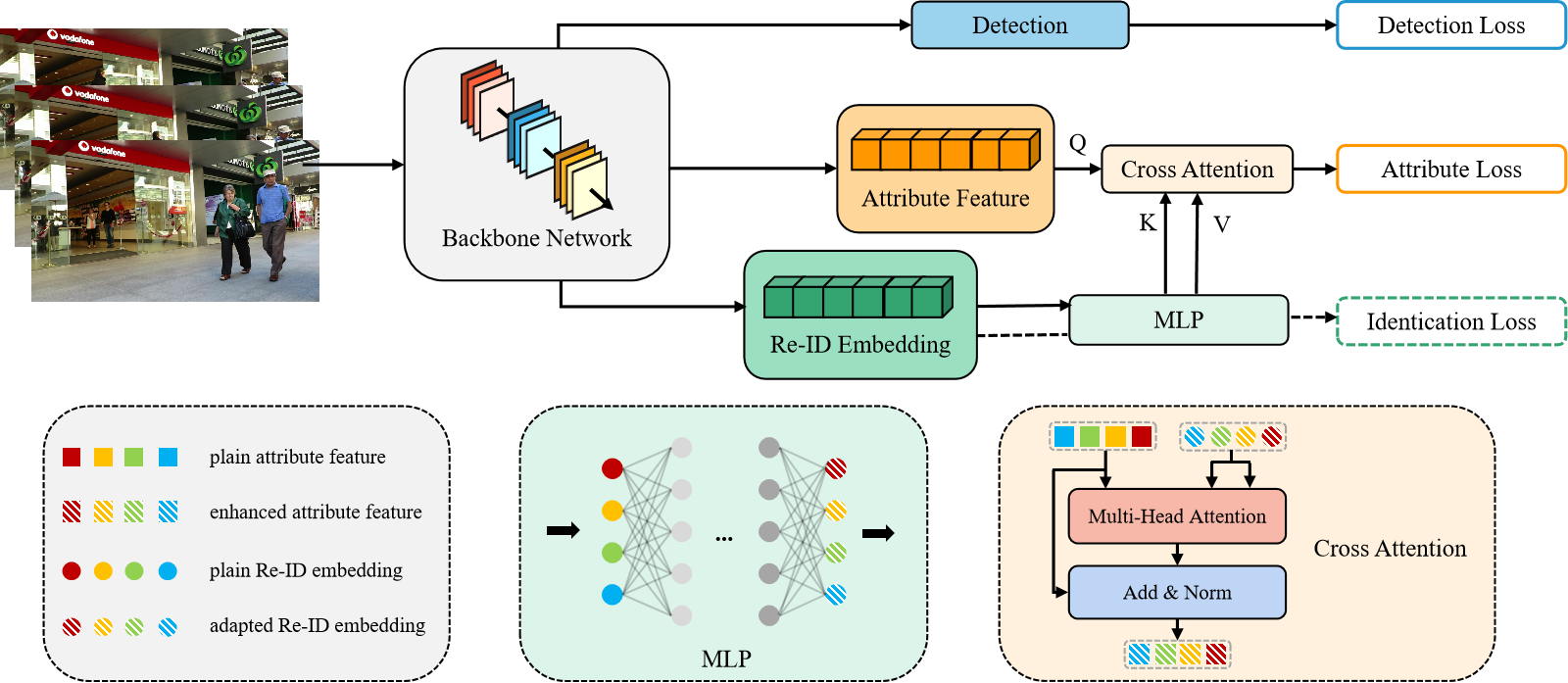}
	\caption{Overall pipeline of the Attribute-Assisted Method (AAM). Given an input pedestrian, we use the original extractor of the implemented tracker, like CNN-based extractor in DeepSORT and Encoder-Decoder structure in FairMOT, to obtain both identification features and attribute features. Subsequently, we use the cross-attention mechanism to enhance the acquisition of attribute features and calculate the attribute loss by the attribute predictions and ground truth labels. As for the identity classification task, we use an identification loss for training and take the attribute predictions as additional aids when testing.}
	\label{fig:5}
\end{figure*}

\subsection{Proposed Method}
Let $P_I = \left\{(x_1, y_1), (x_2, y_2), ..., (x_n, y_n)\right\}$ be a regular identity-labeled pedestrian tracking dataset, where $x_i$ and $y_i$ represent the ~\textit{i}-th pedestrian and its identity label, respectively. Different from normal pedestrian tracking dataset, for each pedestrian $x_i \in P_I$, we provide a multidimensional attribute annotation $a_i = (a_i^1, a_i^2, a_i^3, ..., a_i^m)$, where $a_i^j$ means the ~\textit{j}-th attribute label of the ~\textit{i}-th pedestrian, and ~\textit{m} is the number of the defined attribute classes, which in our work is 32. Now we can let $P_A = {(x_1, a_i), (x_2, a_2), ..., (x_n, a_n)}$ be the attribute labeled dataset. Based on these two sets $P_I$ and $P_A$, we first discuss training strategy and then demonstrate two baselines.

\subsubsection{Training Strategy}
To use attributes to assist pedestrian tracking, we need to extract both Re-ID embedding and attribute vectors from the pedestrian image. For regular Re-ID embedding or ID-discriminative embedding, following the training strategy of most prior works like FairMOT~\cite{zhang2021fairmot}, we regard its training process as a pedestrian identity classification task. For this we present the following function: 
\begin{equation} \label{eq1}
    \sum_{i=1}^n \mathcal{L}(\mathcal{F_I}(w_I^2;\mathcal{E_I}(w_I^1;x_i)), y_i),
\end{equation}
where $\mathcal{E_I}$ is the embedding extraction function, parameterized by $w_I^1$, and $\mathcal{F_I}$ is an identity classifier, parameterized by $w_I^2$. $\mathcal{F_I}$ classify the extracted and embedded pedestrian into a k-dimension identity confidence estimation, in which k is the number of identities. $\mathcal{L}$ represents the loss function. As for training to get a satisfactory attribute vector, we use the strategy proposed by prior pedestrian attribute recognition work~\cite{li2015multi}, as shown in the following objective function:
\begin{equation} \label{eq2}
    \sum_{i=1}^n \mathcal{L}(\mathcal{E_A}(w_A;x_i), a_i),
\end{equation}
where $\mathcal{E_A}$ is the attributes extraction function, parameterized by $w_A$ and $\mathcal{L}$ represents the weighted binary cross-entropy loss function.

\subsubsection{baseline 1}
Pedestrian tracking is a crucial task in computer vision that requires accurately associating pedestrians across multiple frames. One of the main challenges in pedestrian tracking is dealing with occlusions and similar appearances. To address this challenge, in this paper we first propose a simple yet effective baseline approach that combines appearance-based features, such as Re-ID embeddings, with attribute-based features, such as body shape and clothing color.
For trackers that have separate detection and association models, An extra ResNet50 is deployed to extract attribute vectors and concatenate them with the corresponding Re-ID embeddings to obtain an attribute-assisted Re-ID embedding. This approach is surprisingly effective and can improve pedestrian tracking performance.
For trackers that use a joint detection and association model, an attribute head is added to extract pedestrian attributes while also extracting regular Re-ID embeddings. In this case, I compute two distance matrices for the two types of features (Re-ID embeddings and attributes) and then add them together to complete the matching operation.

\subsubsection{baseline 2}
The second method proposed in this paper is the Attribute-Assisted Method (AAM), which aims to further improve the accuracy of pedestrian tracking. To achieve this, AAM incorporates the attention mechanism, which has been shown to be effective in various computer vision tasks, to obtain better attribute vectors.

The pipeline of our proposed AAM is shown in Figure~\ref{fig:5}. It consists of three main parts: the feature extractor, the adaptor, and a cross-attention module. The feature extractor extracts features from the input data, such as images or videos, while the adaptor is responsible for processing these features and generating attribute vectors. The cross-attention module then uses these attribute vectors to improve the accuracy of pedestrian tracking. AAM is generally applicable to various existing pedestrian tracking methods. For example, when applying AAM to the FairMOT algorithm proposed by Zhang et al.~\cite{zhang2021fairmot}, we adopt its original encoder-decoder approach and add an additional feature extraction head to obtain feature predictions. To protect the original extraction ability of the Re-ID embedding extraction head, assuming that it was well-trained beforehand, we fixed its weights and did not update them during the training process in this work. By incorporating the attention mechanism and our proposed AAM method, we can further improve the accuracy of pedestrian tracking and address some of the limitations of existing approaches.

Here is a more detailed explanation of the individual components of the Attribute-Assisted Method (AAM) and how they work together to improve pedestrian tracking accuracy.An adaptor network is designed after obtaining the Re-ID embedding by the backbone network. The adaptor is a MLP block with a residual connection, as shown in the following equation:
\begin{equation} \label{eq3}
   E_2 = Relu(\mathcal{L}_2(\mathcal{L}_1(E_1; w_1); w_2)) + E_1, 
\end{equation}
where $w_1$ and $w_2$ are linear transformation weights of $\mathcal{L}_1$ and $\mathcal{L}_2$ in the MLP module, $E_1$ represents the original Re-ID embedding while $E_2$ is the adapted one. After the adaptor, a feature fusion module is designed to fuse Re-ID features and attribute features. It applies a cross-attention mechanism to the Re-ID embeddings generated by the feature extraction head and the feature predictions generated by the additional feature extraction head and it takes the adapted Re-ID feature as the key and the value, and use the corresponding semantic attributes as the query. The cross-attention structure can be formulated as:
\begin{equation} \label{eq4}
   A_2 = CrossAttn(K=E_2, V=E_2, Q=A_1), 
\end{equation}
\begin{equation} \label{eq5}
   CrossAttn = Softmax((W_\mathcal{Q}Q)(W_\mathcal{K}K)^\top)W_\mathcal{V}V, 
\end{equation}
where $E_2$ represents the adapted Re-ID embedding, $A_1$ is the original attribute feature and $A_2$ is the adapted one. $W_\mathcal{Q}$, $W_\mathcal{K}$, and $W_\mathcal{V}$ are the weights of three attention matrices. This improves the association of pedestrians across frames by incorporating both appearance-based features and attribute-based features. The output of the cross-attention module is our final attribute predictions, when training, we then use the aforementioned weighted binary cross-entropy loss (Equation~\ref{eq2} function to get better attributes, and when testing, methods like vector concatenation and distance matrix addition are adopted to assist pedestrian association.

\begin{figure}[t]
	\centering
	\vspace{2mm}
	\includegraphics[width=0.98\linewidth]{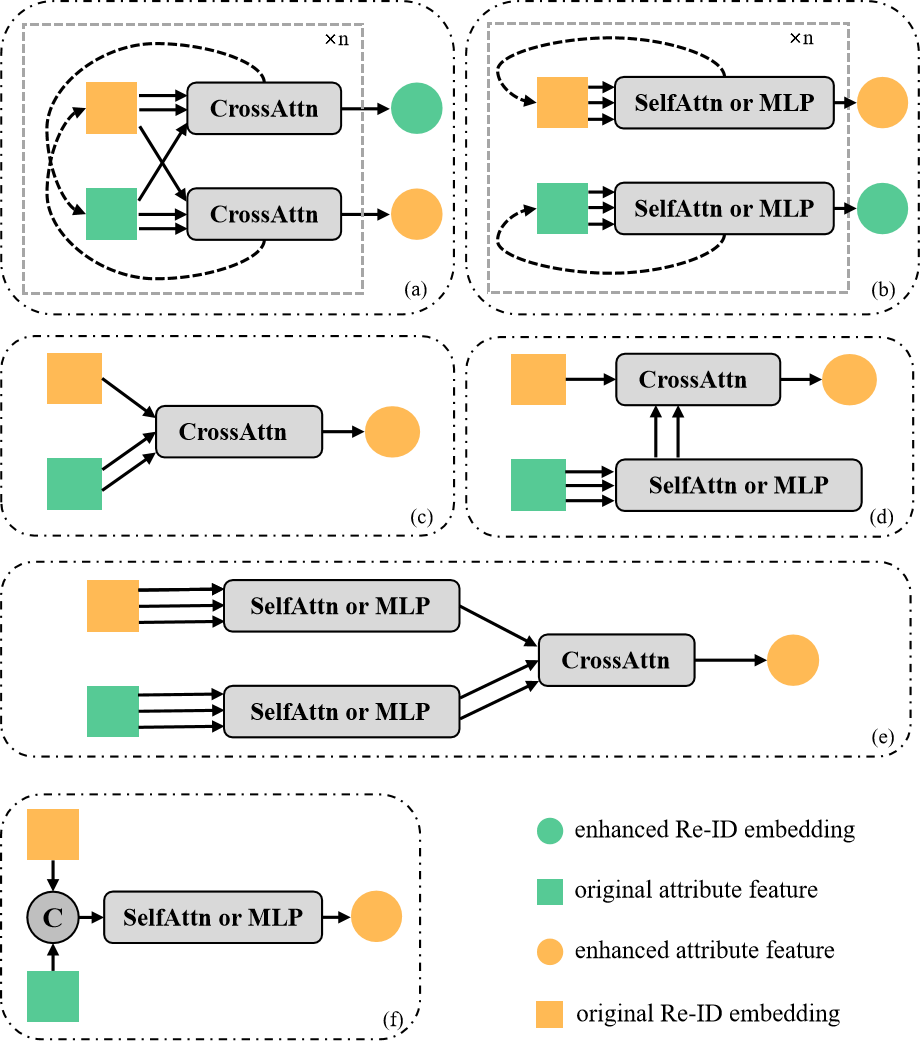}
	\vspace{-1mm}
	\caption{Analysis of various feature fusion strategies. We present some explorations in the area of strategies for fusing semantic attributes and general Re-ID embedding.}
	\label{fig:6}
	\vspace{-1mm}
\end{figure}

\subsection{Analysis of Feature Fusion}
In order to make better use of semantic attributes, we conduct several experiments on different fusion strategies and hope that these attempts will help the future development of feature fusion. Further efforts are desired to design appropriate strategies for the integration of semantic attributes and general Re-ID embedding. Our attempted approaches are illustrated in Figure~\ref{fig:6}.

We first use two cross-attention modules to perform cross-fertilization of semantic attributes and general Re-ID embedding. As shown in Figure~\ref{fig:6}(a), such a process can be repeated any number of times, and the enhanced attributes and embedding can be fused again until the best results are obtained. Strictly, the second method is a self-enhancement method rather than a fusion strategy. As Figure~\ref{fig:6}(b) shows, we use two self-enhancement modules to enhance both features, and just like the first one, such a self-enhancement process can be repeated without any limitations. However, after a series of experiments, we found that these methods are not very effective. We then design four approaches that integrate only attributes into the general embedding without cross-fertilization. In Figure~\ref{fig:6}(c), we directly remove the part where Re-ID embedding is incorporated into semantic attributes, like in Figure~\ref{fig:6}(d), we add a self-attention or MLP module to pre-process the semantic attributes before the fusion part, which is the one used in AAM, and like Figure~\ref{fig:6}(e), we additionally add another self-attention or MLP module to pre-process the Re-ID embedding. To simplify the model even further, we design a structure as shown in Figure~\ref{fig:6}(f). We first concatenate semantic attributes and general Re-ID embedding, and then send the concatenated feature to a self-attention or MLP module, where interaction occurs between different bits of these features, thus achieving the effect of integration. Finally, it should be noted that the fusion strategy used by AAM is the one shown in Figure~\ref{fig:6}(d).

Our exploration of fusion strategies is still not comprehensive enough, and the current fusion results are not entirely satisfactory. Further research is needed in this area. By conducting these analyses, we intended to share our experience and hope that it can guide researchers to pay more attention to feature fusion strategies between semantic attributes and Re-ID embedding.

\section{Experiments} \label{Experiments}
In this section, we first describe the implementation details and training strategies of our proposed new attribute-assisted baselines. Then, we verify the effectiveness of AttMOT step by step by comparing the experiment with some real datasets on various tasks. Furthermore, we  apply our dataset and methods to an identity classification task. Finally, comparison experiments with existing state-of-the-art approaches are conducted on standard benchmark datasets, including MOT17~\cite{milan2016mot16} and MOT20~\cite{dendorfer2020mot20}, which successfully demonstrate the applicability of AAM.

\begin{table*}[t]
	\centering
	\renewcommand{\arraystretch}{1.2}
	\tabcolsep=2.8mm
	\caption{Comparison with real-world dataset trained trackers on pedestrian multi-object tracking. We use three trackers: DeepSORT, StrongSORT, FairMOT, and CSTrack to experiment on pedestrian tracking tasks and all the evaluations are conducted on MOT20 training set. 'MOT17-mix' is a large dataset consisting of MOT17 training set, caltech pedestrian dataset, citypersons, and crowdhuman. 'Ours' represents our AttMOT datasets. }
	\begin{tabular}{c|c|c|ccccccc}
		\toprule[1.2pt]
		& & & MOTA$\uparrow$ & FN$\downarrow$ & FP$\downarrow$ & IDs$\downarrow$ & IDR$\uparrow$ & IDP$\uparrow$ & IDF1$\uparrow$ \\
		\midrule[0.8pt]
		\multirow{8}{*}{MOT20} & \multirow{2}{*}{DeepSORT} & COCO+Market & 51.4 & \textbf{637224} & \textbf{87479} & 2168 & 52.8 & 90.9 & 66.8 \\
            & & Ours & \textbf{59.2} & 640156 & 87595 & \textbf{1750} & \textbf{62.5} & \textbf{92.7} & \textbf{74.7} \\
            \cline{2-10}
            & \multirow{2}{*}{StrongSORT} & COCO+Market & 42.6 & 642879 & \textbf{4570} & 4151 & 38.1 & 87.1 & 53.0 \\
            & & Ours & \textbf{50.3} & \textbf{554151} & 6208 & \textbf{3583} & \textbf{46.6} & \textbf{90.1} & \textbf{61.4} \\
            \cline{2-10}
            & \multirow{2}{*}{FairMOT} & MOT17-mix & 49.2 & \textbf{418895} & 146091 & 11312 & 43.8 & 62.6 & 48.6 \\
            & & Ours & \textbf{51.8} & 439109 & \textbf{97814} & \textbf{10107} & \textbf{43.8}	& \textbf{62.6} & \textbf{51.5} \\
            \cline{2-10}
            & \multirow{2}{*}{CSTrack} & MOT17-mix & 55.2 & 485656 & 14520 & \textbf{8232} & 40.2 & \textbf{68.7} & 50.7 \\
            & & Ours & \textbf{59.3} & \textbf{445098} & \textbf{8801} & 8520 & \textbf{41.4} & 67.3 & \textbf{51.3} \\
 		\bottomrule[1.2pt]
	\end{tabular}
	\label{tab:3}
\end{table*}

\subsection{Implementation Details} \label{Implementation Detail}
Note that all our experiments are performed in the MOTChallenge evaluation suite. The MOT17 and MOT20 datasets are used for the detection, classification, and especially tracking of pedestrians. To provide clarification, it should be noted that the terms "training set" and "testing set" used in this paper specifically refer to the training and testing sets as provided on the official website. It should be emphasized that no further partitioning of the datasets was conducted during our experiments.

\begin{table}[t]
	\centering
	\renewcommand{\arraystretch}{1.2}
	\tabcolsep=2.8mm
	\caption{Comparison with several real-world datasets on pedestrian detection task. We use precision, recall, and mean average precision (mAP) to evaluate the results. 'Caltech' stands for Caltech Pedestrian dataset, 'MOT20' and 'MOT17' represents their training set.}
	\begin{tabular}{c|c|c|ccc}
		\toprule[1.2pt]
		& & & Precision$\uparrow$  &  Recall$\uparrow$  & mAP$\uparrow$\\
		\midrule[0.8pt]
		\multirow{5}{*}{MOT17} & \multirow{3}{*}{YOLOv5} & COCO & 0.605 & \textbf{0.419} & 0.433 \\
            & & Caltech & 0.615 & 0.320 & 0.384 \\
            & & Ours & \textbf{0.802} & 0.404 & \textbf{0.515} \\
            \cline{2-6}
            & \multirow{2}{*}{JDE} & MOT20 & 0.459 & 0.417 & 0.301 \\
            & & Ours & \textbf{0.512} & \textbf{0.423} & \textbf{0.360} \\
            \midrule[0.8pt]
            \multirow{5}{*}{MOT20} & \multirow{3}{*}{YOLOv5} & COCO & 0.539 & 0.357 & 0.400 \\
            & & Caltech & 0.147 & 0.208 & 0.136 \\
            & & Ours & \textbf{0.737} & \textbf{0.430} & \textbf{0.572} \\
            \cline{2-6}
            & \multirow{2}{*}{JDE} & MOT17 & \textbf{0.641} & 0.424 & 0.361 \\
            & & Ours & 0.509 & \textbf{0.526} & \textbf{0.423} \\
 		\bottomrule[1.2pt]
	\end{tabular}
	\label{tab:2}
\end{table}

We mainly focus on two questions:\textit{ how good is the performance of the models trained with our synthetic dataset in the real scenario of MOTChallenge, and how helpful and universally applicable is our attribute-assisted method?} Therefore, a tightly controlled study is performed using many large-scale datasets such as the COCO dataset~\cite{lin2014microsoft}, CrowdHuman~\cite{shao2018crowdhuman}, and CityPerson~\cite{zhang2017citypersons} to train pedestrian tracking models. We evaluate them on MOT17 and MOT20 using the CLEAR metric~\cite{bernardin2008evaluating}, HOTA metric~\cite{luiten2021hota}, and IDF1~\cite{ristani2016performance}. As for the classification task, we use the real-world MOT dataset MOT17. Following JDE~\cite{wang2020towards} and FairMOT~\cite{zhang2021fairmot}. True Positive Rate (TPR) at a certain False Acceptance Rate (FAR) is employed to assess the identity classification ability of trackers on MOT17.

In the experiments with two-step trackers, ResNet-50~\cite{he2016deep} is adopted as the backbone of the feature extractor, and the network is initialized by ImageNet~\cite{russakovsky2015imagenet} pre-trained models. Taking DeepSORT~\cite{wojke2017simple} as an example, we add a 32-dim fully connected layer followed by batch normalization, a drop-out layer with a default drop rate of 0.5, to the ResNet-50 backbone. The 32-dim (for our AttMOT) attribute vector extracted by this ResNet-based feature extractor is concatenated with the 512-dim original Re-ID embedding provided by DeepSORT, and the 544-dim (512 + 32) feature is then used for identity classification. As far as the one-step tracker is concerned, we have preserved its original structure to the greatest extent possible and have only added an extra head for the feature extraction. 

\subsection{Validation Study on AttMOT}
In this section, verification experiments of the effectiveness and robustness of our proposed AttMOT are conducted. We perform the evaluations with several classic or state-of-the-art trackers, including DeepSORT~\cite{wojke2017simple}, StrongSORT~\cite{du2023strongsort}, JDE~\cite{wang2020towards}, FairMOT~\cite{zhang2021fairmot} and CSTrack~\cite{liang2022rethinking}. We evaluate all these models on the widely used pedestrian tracking datasets MOT17~\cite{milan2016mot16} and MOT20~\cite{dendorfer2020mot20}. 

\subsubsection{Comparison of the synthetic dataset and real-world datasets}
First, we would like to prove the validity of our synthetic data set without using its attributes, which means that we intend to verify that our virtual data can replace or even outperform the real-world data to a certain extent. We report our results on pedestrian detection and tracking in Table~\ref{tab:2} and Table~\ref{tab:3}, respectively. In the detection part, we use the Caltech Pedestrian dataset~\cite{dollar2009pedestrian} (we use 'Caltech' for short in the table), the COCO dataset~\cite{lin2014microsoft}, MOT17, MOT20, and our AttMOT to train YOLOv5 and JDE (only for detection, can be broadly considered as a YOLOv3~\cite{redmon2018yolov3}). As shown in Table~\ref{tab:2}, YOLOv5 and JDE achieve competitive and partially much better results on both MOT17 and MOT20 when trained on our synthetic dataset. For example, YOLOv5 trained on AttMOT outperforms its COCO-trained and Caltech pedestrian trained versions by +0.078 and +0.131 in terms of mAP, respectively. 

Similarly, experiments on pedestrian multi-object tracking task are conducted. We choose DeepSORT~\cite{wojke2017simple}, StrongSORT~\cite{du2023strongsort}, CSTrack~\cite{liang2022rethinking}, and FairMOT~\cite{zhang2021fairmot}. FairMOT and CSTrack are trained on the MOT17-mix dataset (MOT17 training set + Caltech Pedestrian dataset + CityPerson + Crowdhuman), while the detection models of DeepSORT and StrongSORT we use are trained on the COCO dataset~\cite{lin2014microsoft} and the association models are trained on the Market1501 dataset. As shown in Table~\ref{tab:3}, we achieve significantly better results for almost all metrics and trackers. We get +7.8 MOTA and +7.9 IDF1 improvement on DeepSORT, +7.7 MOTA, and +8.4 IDF1 on StrongSORT, over the COCO-trained model. When it comes to state-of-the-art, we also gain +2.6 MOTA and +2.9 IDF1 on FairMOT, +4.1 MOTA, and +0.6 IDF1 on CSTrack, in comparison to their MOT17-mix trained version. 

The comparison group in the experiment used a training set composed of multiple real pedestrian datasets, which actually contains a very large amount of data. Indeed, we acknowledge that there are ways to obtain better results by adding more real datasets into the mixed training set, such as ETHZ~\cite{eth_biwi_00534} and CUHK-SYSU pedestrian dataset~\cite{xiao2016end}. But after comparison experiments using the large MOT17-mix dataset (over 100k frames) trained model, we think these results have been able to demonstrate that our proposed synthetic dataset, AttMOT, can help and replace the real dataset to a considerable extent. It is worth mentioning that the training set in the comparison experiment is only slightly smaller in quantity than the one extracted from AttMOT, and there is no significant difference in quantity between them. This further proves that the effectiveness of AttMOT is due to the assistance of semantic attributes rather than the quantity of images.

\begin{table}[t]
	\centering
	\renewcommand{\arraystretch}{1.1}
	\tabcolsep=2.8mm
	\caption{Comparison with representative pedestrian attribute recognition datasets on DeepSORT. 'Plain' means standard DeepSORT without an attribute extractor, and the experiment is conducted on MOT17 training set.}
	\begin{tabular}{c|c|ccc}
		\toprule[1.2pt]
		& & MOTA$\uparrow$ & IDs$\downarrow$ & IDF1$\uparrow$\\
		\midrule[0.8pt]
		\multirow{5}{*}{MOT17} & Plain & 61.8 & 5558 & 62.8 \\
            & PA100K & 62.2 & 5050 & 63.5 \\
            & Market1501-attribute & 62.3 & 4530 & 63.3 \\
            & DukeMTMC-attribute & 62.1 & 5092 & 63.4 \\
            & Ours & \textbf{62.9} & \textbf{4332} & \textbf{68.7} \\
 		\bottomrule[1.2pt]
	\end{tabular}
	\label{tab:4}
\end{table}

\subsubsection{Comparison with real-world datasets with pedestrian attributes}
As previously mentioned, to the best of our knowledge, AttMOT is the first pedestrian tracking dataset with attribute labels. Existing pedestrian attribute datasets are in the form of non-video sequences and lack IDs. It is not feasible to employ single-stage models for training with these datasets as it would disrupt the existing end-to-end training approach. Therefore, for trackers that use a joint detection and association model, such as CenterTrack~\cite{zhou2020tracking}, AttMOT is the only dataset available for utilizing semantic attributes to assist in the MOT task. However, the situation is different for models like DeepSORT~\cite{wojke2017simple}, which use separate models for pedestrian target detection and feature extraction, respectively.

For instance, in DeepSORT, any pedestrian detection dataset can be used to train its detection model, and the COCO dataset\cite{lin2014microsoft} or Person Re-ID datasets such as Market1501\cite{zheng2015scalable} can be used for feature extraction. In this case, similar to previous work on using semantic attributes to support Re-ID, datasets designed for Pedestrian Attribute Recognition (PAR) can be employed to train an additional attribute extractor, which in our work is a ResNet-50~\cite{he2016deep}.

To assess the effectiveness of our dataset in terms of attribute annotation, we perform a comparative experiment with several existing pedestrian attribute datasets, including PA100K~\cite{liu2017hydraplus}, Market1501-attribute, and DukeMTMC-attribute~\cite{lin2019improving}. PA100K is currently the largest pedestrian attribute recognition dataset, comprising of 100k pedestrian images with 26-dimensional attribute annotations. The Market1501-attribute and DukeMTMC-attribute datasets are derived from Market1501~\cite{zheng2015scalable} and DukeMTMC~\cite{zheng2017unlabeled}, two widely used person re-identification datasets, by manually labeling 27-dimensional attributes for Market1501-attribute and 24-dimensional attributes for DukeMTMC-attribute. The DukeMTMC-attribute dataset contains over 36k images of 1812 people, while the Market1501-attribute dataset contains about 33k images of 1501 pedestrians. These three datasets are currently the most effective and representative datasets for pedestrian attribute recognition and attribute-assisted person re-identification tasks. To evaluate the attribute-assisted tracking performance, we select DeepSORT with the same detection and feature extraction model for maximum control of the variance, and the only difference is the attribute extraction. The experiment is conducted on MOT17~\cite{milan2016mot16},using part of the CLEAR metric~\cite{bernardin2008evaluating} and IDF1~\cite{ristani2016performance}.

\begin{table}[t]
	\centering
	\renewcommand{\arraystretch}{1.1}
	\tabcolsep=2.2mm
	\caption{Identity classification experiments on MOT17 and MOT20 training set. we use the True Positive Rate (TPR) at a certain False Accept Rate (FAR) to evaluate the classification ability. 80k-Synth and 120k-Synth are two sub-datasets of AttMOT, containing about 80k and 120k images. 'MOT17-mix' is explained before as big mixed dataset.}
	\begin{tabular}{c|cccc}
		\toprule[1.2pt]
		& & FAR=0.1 & FAR=0.01 & FAR=0.001\\
		\midrule[0.8pt]
		\multirow{5}{*}{MOT17} & CrowdHuman & 0.657 & 0.264 & 0.094 \\
            & MOT20-mix & 0.803 & 0.394 & \textbf{0.261} \\
            & 80k-Synth & 0.755 & 0.415 & 0.259 \\
            & 120k-Synth & 0.824 & 0.492 & 0.215 \\
            & 120k-Synth+AAM & \textbf{0.837} & \textbf{0.495} & 0.215 \\
            \midrule[0.8pt]
            \multirow{5}{*}{MOT20} & CrowdHuman & 0.475 & 0.177 & 0.062 \\
            & MOT17-mix & 0.564 & 0.233 & 0.097 \\
            & 80k-Synth & 0.561 & 0.229 & 0.096 \\
            & 120k-Synth & 0.635 & 0.270 & 0.099 \\
            & 120k-Synth+AAM & \textbf{0.646} & \textbf{0.295} & \textbf{0.108} \\
 		\bottomrule[1.2pt]
	\end{tabular}
	\label{tab:5}
\end{table}

The results are shown in Table~\ref{tab:4}. It is quite obvious that the model trained by our AttMOT is significantly improved in all three metrics, we get +1.1 MOTA, -22\% ID switch, and +5.9 IDF1 over the standard DeepSORT. When compared to the best results obtained by models trained on these three real-world datasets, we also improve by +0.6 in MOTA and +5.2 in IDF1. Given these consistent improvements, we believe that our AttMOT is much better suited for attribute-assisted MOT and Re-ID than any existing publicly available attribute-annotated dataset.


\begin{table*}[t]
	\centering
	\renewcommand{\arraystretch}{1.2}
	\tabcolsep=2.8mm
	\caption{Comparison experiments of state-of-the-art trackers on training splits of MOT17 and MOT20. For each evaluation metric and each set of experiments, the best results are boldfaced.We apply our AAM to five state-of-the-art trackers, including DeepSORT, StrongSORT, FairMOT, CSTrack and CTracker. 'Synth' stands for our AttMOT dataset, MOT17(20)-mix is the mixed dataset we mentioned before and '+AAM' means we use our attribute-assisted method in that experiment.}
	\begin{tabular}{c|c|cccccccccc}
		\toprule[1.2pt]
		& split & & MOTA$\uparrow$ & FN$\downarrow$ & FP$\downarrow$ & IDs$\downarrow$ & HOTA$\uparrow$ & AssA$\uparrow$ & IDR$\uparrow$ & IDP$\uparrow$ & IDF1$\uparrow$ \\
            \midrule[0.8pt]
		\multirow{19}{*}{MOT17} & \multirow{13}{*}{train} & DeepSORT & 59.6 & \textbf{38066} & 6597 & 670 & 51.8 & 52.0 & 53.9 & 75.8 & 62.6 \\
            & & DeepSORT+AAM & \textbf{60.6} & 38187 & \textbf{6399} & \textbf{605} & \textbf{53.4} & \textbf{55.1} & \textbf{55.9} & \textbf{78.0} & \textbf{65.1} \\
            \cline{3-12}
            & & StrongSORT & 59.5 & \textbf{37928} & 7076 & 433 & 53.0 & 54.0 & 55.7 & 76.8 & 64.6 \\
            & & StrongSORT+AAM & \textbf{60.7} & 39785 & \textbf{4774} & \textbf{429} & \textbf{54.3} & \textbf{57.0} & \textbf{56.0} & \textbf{81.4} & \textbf{66.4} \\
            \cline{3-12}
            & & FairMOT(MOT20-mix) & 59.7 & \textbf{41013} & 11312 & 844 & 44.0 & 46.4 & 60.2 & 85.7 & 70.6 \\
            & & FairMOT(Synth) & 62.3 & 42095 & 8849 & 732 & 45.4 & 48.9 & 61.1 & 89.3 & 73.7 \\
            & & FairMOT(Synth)+AAM & \textbf{63.4} & 41699 & \textbf{8173} & \textbf{606} & \textbf{47.1} & \textbf{51.3} & \textbf{62.5} & \textbf{91.7} & \textbf{75.5} \\
            \cline{3-12}
            & & CSTrack(MOT20-mix) & 65.2 & \textbf{39672} & 6789 & 958 & 53.1 & 54.2 & 59.3 & 86.6 & 71.8 \\
            & & CSTrack(Synth) & 68.2 & 40112 & 4562 & 811 & 54.5 & 56.9 & 60.8 & 88.1 & 74.4 \\
            & & CSTrack(Synth+AAM) & \textbf{69.3} & \textbf{41237} & \textbf{4312} & \textbf{745} & \textbf{56.0} & \textbf{58.0} & \textbf{62.1} & \textbf{91.3} & \textbf{77.2} \\
            \cline{3-12}
            & & CTracker(MOT20-mix) & 64.6 & \textbf{42435} & 7789 & 1240 & 47.8 & 48.4 & 58.2 & 79.0 & 59.3 \\
            & & CTracker(Synth) & 66.2 & 43490 & 5893 & 1173 & 49.3 & 50.6 & 59.7 & 80.8 & 61.6 \\
            & & CTracker(Synth+AAM) & \textbf{67.5} & \textbf{43972} & \textbf{4963} & \textbf{1009} & \textbf{50.8} & \textbf{51.9} & \textbf{61.0} & \textbf{82.0} & \textbf{63.8} \\
            \cline{2-12}
            & \multirow{6}{*}{test} & DeepSORT & 60.4 & \textbf{52255} & 18766 & 1196 & 47.7 & 44.0 & - & - & 58.5\\
            & & DeepSORT+AAM & \textbf{61.6} & 53329 & \textbf{17017} & \textbf{986} & \textbf{50.1} & \textbf{47.3} & - & - & \textbf{60.9} \\
            \cline{3-12}
            & & FairMOT(MOT20-mix) & 60.6 & \textbf{53189} & 21290 & 1278 & 40.3 & 42.2 & - & - & 62.9 \\
            & & FairMOT(Synth+AAM) & \textbf{64.1} & 53321 & \textbf{18760} & \textbf{943} & \textbf{44.2} & \textbf{45.0} & - & - & \textbf{65.6} \\
            \cline{3-12}
            & & CSTrack(MOT20-mix) & 64.9 & \textbf{51298} & 13457 & 1312 & 50.1 & 51.8 & - & - & 59.8 \\
            & & CSTrack(Synth+AAM) & \textbf{67.0} & 52674 & \textbf{11296} & \textbf{1074} & \textbf{52.8} & \textbf{53.7} & - & - & \textbf{64.2} \\
            \midrule[0.8pt]
            \multirow{19}{*}{MOT20} & \multirow{13}{*}{train} & DeepSORT & 51.4 & 503021 & \textbf{28077} & 10600 & 23.4 & 17.5 & 52.8 & 90.9 & 66.8 \\
            & & DeepSORT+AAM & \textbf{59.2} & \textbf{408290} & 35381 & \textbf{9247} & \textbf{30.9} & \textbf{24.7} & \textbf{62.5} & \textbf{92.7} & \textbf{74.7} \\
            \cline{3-12}
            & & StrongSORT & 42.6 & 642879 & \textbf{4570} & 4151 & 28.5 & 27.4 & 38.1 & 87.1 & 53.0 \\
            & & StrongSORT+AAM & \textbf{50.5} & \textbf{563755} & 5351 & \textbf{3203} & \textbf{36.3} & \textbf{35.5} & \textbf{46.6} & \textbf{90.1} & \textbf{61.4} \\
            \cline{3-12}
             & & FairMOT(MOT17-mix) & 49.2 & \textbf{418895} & 146091 & 11312 & 37.7 & 32.7 & 42.9 & 56.3 & 48.6 \\
            & & FairMOT(Synth) & 51.8 & 439109 & \textbf{97814} & 10107 & 39.1 & 35.2 & 43.8 & 62.6 & 51.5 \\
            & & FairMOT(Synth)+AAM & \textbf{52.5} & 431091 & 98204 & \textbf{9850} & \textbf{39.4} & \textbf{35.6} & \textbf{44.0} & \textbf{63.1} & \textbf{51.9} \\
            \cline{3-12}
            & & CSTrack(MOT17-mix) & 56.0 & \textbf{41098} & 12471 & 11095 & 32.2 & 34.5 & 43.1 & 60.1 & 48.8 \\
            & & CSTrack(Synth) & 57.4 & 431225 & 10018 & 8890 & 35.6 & 37.2 & 44.7 & 62.9 & 51.2 \\
            & & CSTrack(Synth+AAM) & \textbf{59.3} & 421180 & \textbf{9971} & \textbf{7651} & \textbf{36.1} & \textbf{38.0} & \textbf{45.1} & \textbf{63.6} & \textbf{52.2} \\
            \cline{3-12}
            & & CTracker(MOT17-mix) & 54.6 & \textbf{51437} & 33568 & 18491 & 27.8 & 28.1 & 44.0 & 58.9 & 51.2 \\
            & & CTracker(Synth) & 55.9 & 52389 & 31479 & 16782 & 29.3 & 30.4 & 44.6 & 62.4 & 54.3 \\
            & & CTracker(Synth+AAM) & \textbf{56.5} & \textbf{54468} & \textbf{28765} & \textbf{15560} & \textbf{31.0} & \textbf{31.3} & \textbf{45.3} & \textbf{63.0} & \textbf{55.4} \\
            \cline{2-12}
            & \multirow{6}{*}{test} & DeepSORT & 50.4 & \textbf{200912} & 50139 & 3310 & 30.7 & 25.1 & - & - & 50.1\\
            & & DeepSORT+AAM & \textbf{53.8} & 231298 & \textbf{6439} & \textbf{2178} & \textbf{32.7} & \textbf{29.0} & - & - & \textbf{56.5} \\
            \cline{3-12}
            & & FariMOT(MOT17-mix) & 49.8 & \textbf{218738} & 35014 & 4509 & 39.7 & 44.0 & - & - & 49.4 \\
            & & FariMOT(Synth+AAM) & \textbf{52.4} & 236579 & \textbf{33178} & 4123 & \textbf{43.0} & \textbf{44.1} & - & - & \textbf{53.1}\\
            \cline{3-12}
            & & CSTrack(MOT17-mix) & 57.1 & \textbf{187437} & 30298 & 1674 & 45.9 & 45.7 & - & - & 57.7 \\
            & & CSTrack(Synth+AAM) & \textbf{60.2} & 202516 & \textbf{24434} & \textbf{1436} & \textbf{48.7} & \textbf{49.0} & - & - & \textbf{61.4} \\
 		\bottomrule[1.2pt]
	\end{tabular}
	\label{tab:6}
\end{table*}

\subsection{Evaluation of identity classification}
Finally, we evaluate the performance of the person re-identification (Re-ID) model to demonstrate further uses of our dataset. We train FairMOT on several real-world datasets and two sub-datasets of our AttMOT, and then evaluate it on the MOTChallenge MOT17 and MOT20 datasets by testing its ability to classify identities.  

As can be seen in Table~\ref{tab:5}, by training on the 120k-Synth dataset, we already outperform models trained on real-world datasets like CrowdHuman, MOT17-mix, and MOT20-mix, e.g. we gain +0.167 in terms of FAR=0.1 for CrowdHuman and +0.034 for MOT20-mix when evaluating on MOT17. When we apply our AAM to the tracker, we observe an overall improvement of +0.013 and +0.011 in terms of FAR=0.1 on MOT17 and MOT20, respectively. This suggests that both of our datasets and methods, AttMOT and AAM, can effectively enhance the identity classification capability of the existing tracker, and are very likely to be applicable to the person re-identification task.

\subsection{Comparison with state-of-the-art trackers}
Our proposed Attribute-Assisted Method (AAM) is evaluated on the pedestrian multi-object tracking datasets MOTChallenge MOT17~\cite{milan2016mot16} and MOT20~\cite{dendorfer2020mot20} using five trackers, namely DeepSORT~\cite{wojke2017simple}, StrongSORT~\cite{du2023strongsort}, FairMOT~\cite{zhang2021fairmot}, CSTrack~\cite{liang2022rethinking}, and CTracker~\cite{peng2020chained}. The performance of AAM is compared against the baseline trackers, and the results are reported in Table~\ref{tab:6}. Our series of experiments and previous studies have shown that finetuning on the MOT17 or MOT20 training sets significantly enhances the performance of the model on the corresponding test sets. To ensure a fair comparison, unlike conventional MOT experiments, we refrained from finetuning the model on the MOT17 or MOT20 training sets while verifying the efficacy of AttMOT. In other words, the training and test datasets for MOT17 or MOT20 used in our comparative experiments comprise images that the model has not been trained on, and therefore hold equal significance. As a result, we conducted validation experiments on both the training and test sets to ensure a thorough evaluation. While the top-performing models on the MOT leaderboard utilized a large amount of training data and conducted finetuning on the corresponding MOT training set, this is not the focus of our study. Our aim in this paper is to conduct a fair comparison with the baseline models to demonstrate the effectiveness of AttMOT and AAM, and prove their assistance to current pedestrian tracking task.

For DeepSORT and StrongSORT, two types of experiments are conducted on each dataset, plain version and attribute-assisted version, to investigate the impact of the semantic attributes on tracking performance. The results show that incorporating the semantic attributes in the tracking model significantly improves the performance of both DeepSORT and StrongSORT, with a boost of up to +8.1 in terms of IDF1 on MOT20 training split. For FairMOT, CSTrack and CTracker, we train four models using different datasets, including MOT17-mix (as explained before, MOT17-mix represents a large-scale dataset composed of the MOT17 training set, Caltech Pedestrian dataset, CityPerson, and Crowdhuman), MOT20-mix, plain AttMOT, and full AttMOT, and the proposed AAM consistently outperforms the baseline models across all metrics. Moreover, even when only considering the attribute part, AAM is able to generate notable improvements in most metrics, such as +1.1 in terms of MOTA, +1.7 in terms of HOTA, and +1.8 in terms of IDF1 when evaluating on MOT17 train split and using the AAM and AttMOT methods helped FairMOT improve its MOTA by 3.5, HOTA by 3.9, and IDF1 by 2.7 on the MOT17 testing set. The results in Table~\ref{tab:6} demonstrate the strong generalization capacity of AAM and its potential to enhance the performance of existing trackers. Validation experiments conducted on both the training and test sets of MOT17 and MOT20 yielded similar results, which further substantiates the effectiveness and generalization ability of AAM and AttMOT.

\begin{figure*}[!t]
	\centering
	\includegraphics[width=0.98\linewidth]{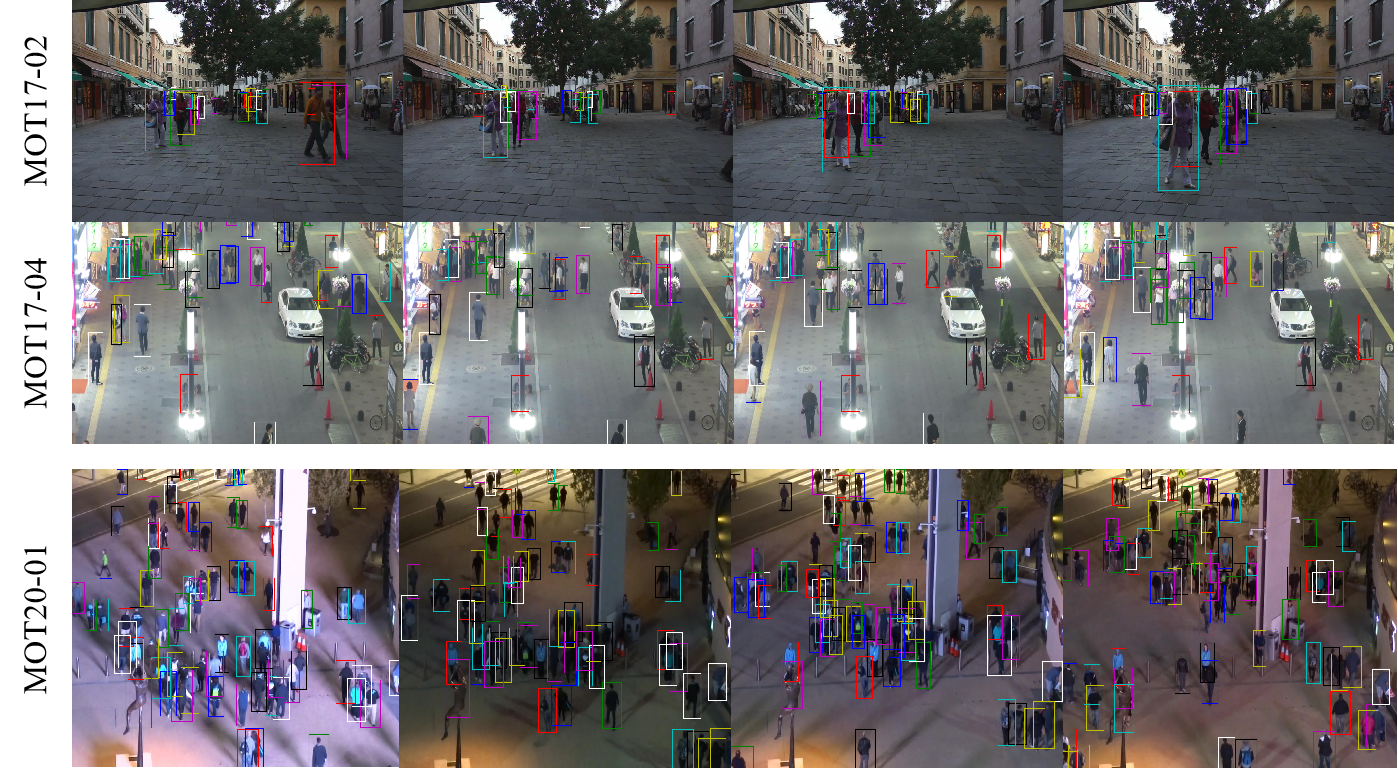}
	\caption{\textbf{Example tracking results of our method on MOT17 and MOT20.}Each row displays the results of the sampled frames in chronological order of a particular video sequence. Bounding boxes are marked in the images with different colors representing different identities. The first two rows of images are from MOT17 and the last one is from MOT20, we can see that our tracking results have good continuity and accuracy on sequences from both datasets.}
	\label{fig:7}
\end{figure*}

Figure~\ref{fig:7} presents visualization results of challenging cases that can be addressed by the attribute-assisted FairMOT model. The results demonstrate that our approach achieves promising performance in terms of detection accuracy, identity classification, and identity continuity.

\subsection{Limitations and challenges}
At present, AAM still faces certain challenges. For instance, in situations where pedestrian images are excessively blurred, pedestrian attribute recognition may become extremely arduous, thereby limiting the assistance that AAM can provide for pedestrian tracking. Moreover, it is noteworthy that both AAM and AttMOT are currently restricted to pedestrian tracking, and similar methodologies need to be extended to track other categories of objects, such as vehicles.

Furthermore, it should be emphasized that the use of semantic attributes to facilitate multi-object tracking has certain inherent limitations. Specifically, in scenarios where a large number of objects share identical or highly similar attributes, such as all players wearing the same uniform on a football field, the effectiveness of semantic-assisted tracking may be limited or even counterproductive. These challenges and limitations warrant further exploration and investigation in future research. Nevertheless, we hold a strong belief that this approach and methodology possess significant potential.

\section{Conclusion}
This paper introduces AttMOT, a novel large-scale synthetic dataset for pedestrian tracking that includes pedestrian attribute annotations, which to the best of our knowledge is the first of its kind. Additionally, we propose a simple yet effective method for attribute-assisted multi-object tracking, called AAM, which can be easily integrated into existing trackers and consistently improves performance. Our experimental results demonstrate that AttMOT and AAM significantly enhance pedestrian detection, classification, and tracking. Notably, applying AAM to current state-of-the-art trackers yields top-performing results on the MOTchallenge benchmark datasets. Moreover, we conduct detailed analysis and exploration of how to better integrate and apply semantic attributes in MOT tasks, which we believe will inspire future research in this area. Overall, we expect that this paper will motivate further exploration of the potential uses of semantic attributes, and we plan to explore the potential of synthetic datasets with semantic attribute annotations in future work.


\vfill

\end{document}